%% file: main.tex
\documentclass[journal,10pt,letterpaper]{IEEEtran}
\usepackage{amsmath,amssymb,amsfonts}
\usepackage{booktabs,multirow,graphicx,cite,url,placeins}
\usepackage[dvipsnames]{xcolor}
\usepackage[hidelinks]{hyperref}
\graphicspath{{figures/}}
\newcommand{\ours}{Mamba3R}
\newcommand{\state}{\mathbf M}
\begin{document}
\title{Mamba3R: Fixed-Capacity Selective State Memory for Long-Sequence 3D Reconstruction}
\author{Zhenxiang Xiong, Tianchen Deng, Nailin Wang, Lei Zhao, Guole Shen, and Weidong Chen%
\thanks{This work has been submitted to the IEEE for possible publication. Copyright may be transferred without notice, after which this version may no longer be accessible.}%
\thanks{This work was supported by the National Natural Science Foundation of China under Grant 62573287. (Corresponding authors: Tianchen Deng; Weidong Chen.)}%
\thanks{The authors are with the Institute of Medical Robotics and the School of Automation and Intelligent Sensing, Shanghai Jiao Tong University, Shanghai 200240, China, and also with the Key Laboratory of System Control and Information Processing, Ministry of Education, Shanghai 200240, China (e-mail: azuresea@sjtu.edu.cn; dengtianchen@sjtu.edu.cn; wnailinxNeil@sjtu.edu.cn; zhao-lei@sjtu.edu.cn; mikrokosmos@sjtu.edu.cn; wdchen@sjtu.edu.cn).}}
\maketitle

\begin{abstract}
Streaming 3D reconstruction must maintain geometric quality over long inputs within a fixed GPU budget. A pretrained recurrent model provides bounded state, but freezing its backbone also fixes its learned state--observation interactions. We introduce \ours{}, a jointly designed state--memory adapter that provides a trainable adaptation path without updating the backbone or task heads. A selective state-space mixer jointly transforms native state features and a fixed-capacity auxiliary memory, explicit read and write paths carry information across frames, and a decomposed feedback path regulates what the added memory feeds back. Only 1.28 million adapter parameters are trained. Under a shared periodic-reset protocol, it lowers trajectory error and improves accumulated surface and depth accuracy over recurrent baselines on public indoor and outdoor datasets and robot-recorded traversals, gaining up to 40 percent on one metric. Matched comparisons support both the choice of the mixing operator and the utility of carrying the auxiliary memory across frames. Allocated memory remains unchanged and throughput moves by at most 1.4 percent from one to two thousand frames. Replaying its recorded traversals offline, the platform's RTX 4060 averages 11.7 frames per second within 5.1~GiB with the same checkpoint, a mean throughput covering the 10~Hz sensor rate.
\end{abstract}

\noindent\textbf{Note to Practitioners---}
Mobile mapping and inspection systems need coherent geometry along a moving camera while sharing the robot's GPU with the rest of its software. This work upgrades an RGB-based recurrent reconstruction model with a compact trainable adapter and auxiliary memory, leaving the inherited backbone and task heads untouched; the module keeps the depth, pose, and point-map interface and drops into an existing perception pipeline. Because the added state has fixed capacity, throughput and memory do not drift with traversal length: a portable implementation replays complete pre-recorded sequences offline on an RTX 4060 at a mean 11.7 frames per second within 5.1~GiB, a throughput covering the platform's 10~Hz camera rate. Practitioners should keep the attenuated memory feedback and history decay reported here, and verify sensor synchronization and calibration before moving from recorded inputs to a live stream.


\input{sections/01_introduction}
\input{sections/02_related_work}
\input{sections/03_method}
\input{sections/04_protocol}
\input{sections/05_results}
\input{sections/06_deployment}

\input{sections/07_mapping}
\input{sections/08_conclusion}

\section*{Acknowledgment}
An AI assistant (Claude, Anthropic) was used for language polishing and editing of the manuscript text, for fixing bugs in parts of the supporting code, for completing parts of the plotting scripts, and for assisting in the aggregation and generation of some of the line charts and tables; the data and the measured numbers are not AI-generated. The authors reviewed and edited all AI-assisted content and take full responsibility for the content of this article.

\bibliographystyle{IEEEtran}
\bibliography{references}
\end{document}

%% file: sections/01_introduction.tex
\section{Introduction}
A mobile platform that inspects a facility or carries a passenger through a building has to build its geometry while it moves, and what the rest of the stack consumes is the quality of that geometry: a small camera-pose error misregisters otherwise plausible local structure, and inconsistent depth deforms the accumulated map. This paper improves that quality with a compact adapter added to a frozen pretrained recurrent reconstruction model, lowering trajectory error and improving accumulated surface and depth accuracy on public benchmarks and on traversals recorded by a wheelchair platform. The second property is what lets the first be used: the adapter's state has fixed capacity, so throughput and memory do not grow with traversal length and the module holds a constant footprint inside a compute budget shared with localization, planning, and control.

Two families of streaming predictor precede this work. Causal-attention models such as StreamVGGT and STream3R commit to a frame on arrival but cache historical keys and values~\cite{zhuo2026streamvggt,lan2026stream3r}, so the retained state grows with the traversal; recurrent models---CUT3R and the TTT3R update rule refined on it---keep a summary of fixed size~\cite{wang2025cut3r,chen2025ttt3r}, the operating point a shared onboard budget requires. Bounded state alone does not prevent drift, motivating periodic resets with predicted-camera bridging. We therefore ask: under a fixed recurrent-state budget and a shared periodic-reset protocol, can a compact trainable adapter improve the geometric quality accumulated over long streams?

Freezing the backbone fixes its learned state--observation transformations; we add a trainable adaptation path beside the inherited predictor. \ours{} is a jointly designed state--memory adapter: a selective state-space mixer transforms native state features and a fixed-capacity auxiliary memory jointly, explicit read and write paths carry information across frames, the inherited per-token confidence update is kept, and a decomposed feedback path with history-only decay regulates the memory contribution (Section~\ref{sec:scannet_ablation} characterizes both interventions). The encoder, decoder, native pose memory, and all geometric heads stay frozen: 1.28 million adapter parameters are trained, and the depth, pose, and point-map interface is unchanged.

The evaluation follows the order a practitioner would ask the questions in: public benchmarks measure accumulated geometry, long-stream trajectory accuracy, and inference cost, with matched comparisons for the mixing operator and the memory carry-over; the robot platform supplies its own recorded traversals and onboard RTX 4060, on which the same checkpoint runs without retuning; finally the streaming predictions are coupled to a fixed loop-closing back-end---offline to isolate the front-end, then incrementally on the device---to test whether a better front-end still produces a better map once that optimizer acts.

\begin{figure*}[!t]
\centering
\includegraphics[width=\textwidth]{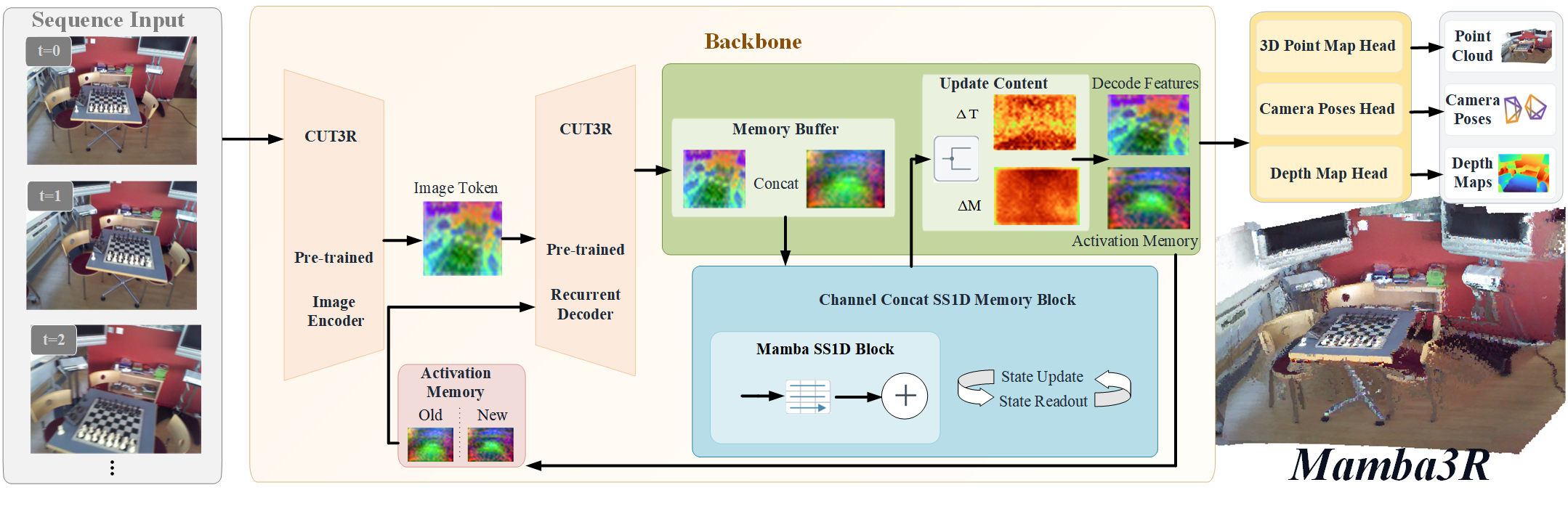}
\caption{Overview of \ours{}. A frozen encoder and recurrent decoder are augmented by the state--memory adapter: it mixes the auxiliary memory with the state carried into the call, its read residual corrects the current frame's decoder feature before the heads run, and its memory half supplies the write read by the next frame (Section~\ref{sec:method}; execution order in supplementary Algorithm~S1).}
\label{fig:method}
\end{figure*}

The contributions separate what is designed from what is measured:
\begin{itemize}
\item \emph{Joint state--memory adaptation on a frozen recurrent backbone.} A trainable selective state-space mixer transforms the native state features and a fixed-capacity auxiliary memory in one pass, with explicit read/write paths carrying information across frames; the inherited encoder, decoder, pose memory, per-token commit rule, and prediction interface are unchanged.
\item \emph{A controlled memory-feedback design.} The adapter's response is decomposed into a zero-memory component and the memory-induced increment, the increment is attenuated, and carried history decays---what the auxiliary memory feeds back is regulated, not injected at full strength.
\item \emph{Experimental validation of both.} With 1.28 million trained parameters, the adapter improves trajectory, accumulated surface, and depth accuracy over the evaluated recurrent comparators under a shared periodic-reset protocol, on public data and robot traversals, at essentially unchanged inference cost; matched comparisons isolate the mixing-operator choice and the memory carry-over, the same checkpoint replays recorded traversals offline on an RTX 4060 at a throughput covering their 10~Hz rate, and the advantage survives a fixed loop-closing back-end, widening on indoor multi-session fusion.
\end{itemize}

%% file: sections/02_related_work.tex
\section{Related Work}
\subsection{Feed-Forward and Streaming Reconstruction}
Learned geometric models replace the correspondence-and-optimization decomposition of structure from motion~\cite{hartley2004multiple,schonberger2016sfm} with direct prediction: DUSt3R regresses pixel-aligned point maps without explicit calibration~\cite{wang2024dust3r}, MASt3R adds a dense matching head~\cite{leroy2024mast3r}, and Fast3R and VGGT scale this to many images with global attention~\cite{yang2025fast3r,wang2025vggt}---at the price that a frame's prediction depends on the whole collection. Streaming reconstruction instead asks the network to commit to a frame on arrival, which turns the design question into one of memory: what to remember, at what cost.

One family keeps the tokens of past frames. In their native continuous mode, StreamVGGT and STream3R make the attention causal and cache historical keys and values~\cite{zhuo2026streamvggt,lan2026stream3r}: nothing is summarized away, so any earlier frame stays individually addressable and the fusion weight is decided per query, but the cache is the history and its size grows with the number of frames, exceeding an embedded budget well before a traversal ends (Section~\ref{sec:results}). Spann3R and Point3R store an external structure instead---a spatial memory for sequential reconstruction and an explicit spatial pointer memory anchored in reconstructed positions~\cite{wang2025spann3r,wu2025point3r}---which bounds the representation by observed scene rather than elapsed time, a genuine improvement, but the state still grows with coverage.

\subsection{Recurrent State and Test-Time Update Rules}
\label{sec:related_state}
A second family retains a summary of fixed size. CUT3R maintains one persistent recurrent representation, updated with each view and read back at a cost independent of sequence length~\cite{wang2025cut3r}; the design question becomes how to overwrite that state. GRS-SLAM3R also uses gated updates to maintain spatial memory~\cite{shen2026grsslam3r}. Test-time training formalizes the forward pass as online learning on an adaptable hidden state~\cite{sun2025ttt}; TTT3R derives a closed-form confidence-dependent update, one rate per state token from state-to-image attention~\cite{chen2025ttt3r}, and ZipMap uses test-time fast weights~\cite{jin2026zipmap}, whose online checkpoint we evaluate with one external frame per call, denoted ZipMap-W1.

Pretrained recurrent models already couple state and image features; freezing their weights preserves these learned interactions while leaving them unavailable for further training. Three inherited components therefore fix the division of labor in this work: CUT3R supplies the native recurrent state and the pretrained state--observation interactions~\cite{wang2025cut3r}; TTT3R supplies the per-token confidence commitment rule, retained unchanged~\cite{chen2025ttt3r}; and Mamba supplies the selective mixing operator~\cite{gu2024mamba}. None of the three provides a trainable path that adapts what the frozen state carries. What this paper adds is the specific adaptation structure assembled on top of them: a joint state--memory mixing interface beside the inherited predictor, explicit read and write paths that carry a fixed-capacity auxiliary memory across frames, and a decomposed, attenuated feedback that controls what that memory contributes. Inference constructs no test-time loss and optimizes no parameters. Mamba makes part of a state-space transition depend on the current input~\cite{gu2024mamba} and VMamba adapts selective scanning to visual representations~\cite{liu2024vmamba}; the property we need is that such a mechanism can mix feature slots within each frame instead of scanning a growing sequence of stored images, so its per-frame cost is fixed. SaLon3R attacks the same growth at a different layer of the same backbone~\cite{guo2026salon3r}, compressing what it emits into quantized Gaussian anchors, so capacity is bounded in the 3D primitive space rather than inside the recurrent state; as neither code nor checkpoints are released and its protocol scores rendering quality, we position our work relative to it without entering its numbers.

\emph{Sibling state-update rules.} Two modifications of this family appeared alongside this work. TTSA3R~\cite{zheng2026ttsa3r} leaves the backbone untouched and replaces only the commitment rule; we evaluate it under our protocol with the accuracy tables. Mem3R~\cite{liu2026mem3r} splits memory between an implicit fast-weight MLP and an explicit token state but replaces the pose state and decoder layers---a different backbone rather than a rule over the same one---and ships no weights at present. LoGeR~\cite{zhang2026loger} is chunk-causal and reports peaks of 18.1--27.2~GB. Mamba-VGGT~\cite{deng2026mambavggt} carries an external memory token across overlapping VGGT windows with a sliding-window selective scan. We do not claim the combination of a selective state-space mechanism with an external geometric memory as new: the shared idea is that a selective scan moves information across a bounded state instead of retaining every past frame. The difference is the object. Mamba-VGGT reasons across temporal windows of a bidirectional backbone whose attention must be truncated, distilling each window forward through a FIFO buffer of patch tokens; this work adapts a frozen recurrent predictor frame by frame, on the constant-size persistent state it already carries, so no window is needed and no frame waits for its successors. Its peak allocation stays bounded at 7.6--9.1~GB over 16--292 frames, and its throughput under the window-update mechanism does not meet the forward requirement of onboard operation, so it cannot serve the goal of this work: under a frozen prediction interface, periodic reset, and an onboard budget, whether complete-trajectory accuracy and downstream reconstruction improve. Our scan also runs one direction (SS1D) rather than four---the CUT3R state adapted here has no patch-grid correspondence, so extra directions have no axis to exploit and are dropped for cost---and periodic reset follows the reported behavior of TTT3R~\cite{chen2025ttt3r} rather than originating here. What these systems do not yet supply is a configuration an onboard computer can hold. TTSA3R's measured configuration does not add tensor memory over the CUT3R recursion it modifies and costs about 16\% of its rate, but it is not entered into the deployment comparison: its timed path omits stages the other rows include, and its best operating point is a different protocol (supplementary material); LoGeR's reported peaks sit above the onboard budget still.

\subsection{Complete SLAM Systems and Optimization Back-Ends}
\label{sec:related_backend}
A streaming predictor is not a complete SLAM system, which integrates pose estimation and scene mapping~\cite{deng2025neslam,deng2025incrementaljoint,deng2026voxelized}; recent work builds such systems on feed-forward geometric priors: MASt3R-SLAM attains real-time dense monocular SLAM from a two-view matching prior~\cite{murai2025mast3rslam}, and VGGT-SLAM incrementally aligns feed-forward submaps over the SL(4) manifold~\cite{maggio2025vggtslam}. Both buy global consistency with repeated matching and optimization whose working set grows with the mapped extent. Our contribution sits at the front-end position such systems occupy, as a constant-cost alternative, and Section~\ref{sec:mapping} measures what happens when its predictions are handed to an optimizing back-end rather than scored open-loop.

Global consistency over a complete traversal is normally recovered by such a back-end, which detects revisits and redistributes residual error. LEMON-Mapping targets large-scale multi-session point-cloud maps~\cite{wang2026lemon}, rejecting outlier loops and propagating refined constraints through a two-step pose-graph optimization; its interface is poses with registered scans, so a learned streaming predictor can take the place of the LiDAR odometry it was built around. No such back-end supplies the loop evidence itself: what it achieves depends on whether the front-end produced geometry consistent enough for a revisit to verify. Section~\ref{sec:mapping} uses exactly that property, holding one optimizer fixed while the front-end is substituted; the optimizer there is an independent RGB-only similarity pose graph following LEMON's organization without bundle adjustment (differences in the supplementary material).

%% file: sections/03_method.tex
\section{Selective Activation Memory}
\label{sec:method}
Each design element below answers one limitation that freezing the backbone creates. Freezing fixes the pretrained state--observation interactions: the selective mixer adds a trainable path without touching the inherited one. The native state offers no location where task-side content can persist without overwriting pretrained content: the slot-aligned, fixed-capacity auxiliary memory with explicit read and write paths supplies it. And a predictor trained without an added signal can be destabilized by one injected at full strength: decomposed feedback and history-only decay regulate what reaches the frozen predictor. The components built on---CUT3R's recurrent state, TTT3R's per-token commitment rule, Mamba's selective scan---are inherited, not contributed (Section~\ref{sec:related_state}); the contribution is the adaptation structure connecting them.

\subsection{Recurrent Geometric Interface}
Figure~\ref{fig:method} summarizes the recurrent interface. Let $I_t$ denote the incoming RGB image. The pretrained encoder--decoder maintains a recurrent scene representation $S_{t-1}\in\mathbb R^{L\times C}$ and a native pose memory $P_{t-1}$. It processes the current observation to produce a candidate scene state $\widetilde S_t$ and the features consumed by the inherited depth, pose, and point-map heads. We introduce a persistent activation memory $\state_{t-1}\in\mathbb R^{L\times K}$, initialized to zero and aligned with the recurrent slots. The evaluated configuration uses $K=64$. Neither the slot count nor the memory width grows with the number of input frames.

The native decoder does not read the added memory. The mixer receives $[\state_{t-1};S_{t-1}]$: the previously stored memory and the state carried into this call, not the decoder's updated feature $\widetilde S_t$.

\subsection{Selective Mixing of State and Memory}
The memory and recurrent features are concatenated along channels (the interface is drawn in the supplementary material):
\begin{equation}
 U_t=[\state_{t-1};S_{t-1}],\qquad
 (V_t^M,V_t^S)=\mathcal B_\theta(U_t).
 \label{eq:joint}
\end{equation}
The mixer preserves the combined width and splits its output into memory-side and state-side features. The indices in~\eqref{eq:joint} carry the temporal semantics: the mixer consumes the state and memory \emph{committed at frame $t-1$}, its state-side output corrects the decoder's updated feature for the \emph{current} frame (Section~\ref{sec:feedback_control}), and its memory-side output becomes the write that frame $t+1$ reads---content crosses frames through two carriers one step apart. A projection and depthwise one-dimensional convolution produce the scan content, while an additional branch supplies a multiplicative gate. For slot index $\ell$, a simplified selective state-space step is
\begin{align}
 h_\ell &= \overline A_\ell h_{\ell-1}+\overline B_\ell x_\ell,\\
 y_\ell &= C_\ell h_\ell+D x_\ell,\qquad
 \overline A_\ell=\exp(\Delta_\ell A).
 \label{eq:ssm}
\end{align}
Here $x_\ell\in\mathbb R^{D_s}$ is slot $\ell$'s projected feature with scan width $D_s$, $h_\ell\in\mathbb R^{D_s\times N}$ holds $N$ internal states per scan channel, and $\Delta_\ell\in\mathbb R^{D_s}$, $\overline B_\ell$, $C_\ell$ are computed from $x_\ell$, so the transition is input-dependent; normalization, gating, and an output projection complete $\mathcal B_\theta$. $A=-\exp(A^{\log})$ is learned, and $\overline A_\ell=\exp(\Delta_\ell A)$ is the zero-order-hold discretization of a continuous-time transition: $\Delta_\ell$ acts as a per-slot step size that sets how strongly slot $\ell$ overwrites the carried scan state, which is the selectivity the design borrows. The scan axis is the slot index: slots are the backbone's state tokens in their fixed inherited order (no spatial or patch-grid meaning, identical at training and inference), with the auxiliary memory aligned one-to-one. A single forward scan is used, so slot $\ell$ conditions only on earlier slots and the computation is not permutation-invariant; the matched lightweight-attention mixer of Section~\ref{sec:scannet_ablation} supplies a permutation-equivariant control without an imposed scan direction, which the forward scan outperforms on all three ScanNet metrics, though no other slot ordering is explored. The scan requires no future frames and no two-dimensional patch grid; $h_\ell$ is internal to the mixer call, whereas $\state_t$ is retained across frames.

\subsection{Controlling Memory-Dependent Feedback}
\label{sec:feedback_control}
Let $r_t$ be the read mask and let $P_S$ be the state-residual projection. Define the adapter response
\begin{equation}
 R_\theta(S,M)=\gamma\,r_t\odot P_S\!\left(\mathcal B_\theta([M;S])_S\right),
 \label{eq:response}
\end{equation}
where $\gamma=0.5$ in the evaluated configuration. Instead of attenuating the entire adapter, we separate the response with zero external memory from the increment induced by memory:
\begin{align}
 R_t^0 &= R_\theta(S_{t-1},0),\\
 \Delta R_t^M &= R_\theta(S_{t-1},\state_{t-1})-R_t^0,\\
 S_t^+ &= \widetilde S_t+R_t^0+\alpha\Delta R_t^M,
 \qquad \alpha=0.5.
 \label{eq:feedback}
\end{align}
The zero-memory branch still receives the carried state, which already incorporates native recurrent history, so the decomposition isolates dependence on the \emph{added} memory; the same mixer weights serve both evaluations, and the extra computation is inside every runtime number we report.

The scene-state commit follows the inherited attention-gated update, i.e., the per-token confidence rescaling that TTT3R introduced. For binary update mask $u_t$ and slot-wise coefficient $g_t$ derived from state-to-image attention---one rate per state slot, shared across that slot's channels---
\begin{equation}
 S_t^{\mathrm{pred}}=(u_t g_t)\odot S_t^+ +(1-u_tg_t)\odot S_{t-1}.
 \label{eq:commit}
\end{equation}
The first-frame branch uses the native initialization rule. The committed state $S_t^{\mathrm{pred}}$ is what the heads of frame $t$ read, so the correction applies on the frame that computes it; normally $S_t=S_t^{\mathrm{pred}}$, both this frame's prediction state and the state carried into the next call, while a reset restores the native initial state for the next call after preserving this frame's prediction.

\subsection{Masked Writes and History-Only Decay}
The memory-side feature is projected into a residual write:
\begin{equation}
 \Delta\state_t=w_t\odot P_M(w_t\odot V_t^M),
 \qquad \widehat{\state}_t=\state_{t-1}+\Delta\state_t,
 \label{eq:write}
\end{equation}
with a binary write mask $w_t$. The outer mask prevents projection bias from updating an unwritten slot. The final stored memory is
\begin{equation}
 \state_t=\rho_t\state_{t-1}+\Delta\state_t,
 \label{eq:leak}
\end{equation}
for a non-reset step. Let $a$ count effective writes in the segment. During the first 32 writes $\rho_t=1$ (writes take effect, history undecayed); later valid writes use $\rho_t=0.8^{1/12}\approx0.9816$; a masked write keeps $\rho_t=1$ and preserves the native candidate---equal to the incoming memory under~\eqref{eq:write}---advancing neither $a$ nor decay.

At a reset, the stored scene and pose states return to their native initial values, the memory is zero, and $a=0$; no old history leaks across the reset. The reset flag belongs to the original, scored boundary frame; its duplicate (reset false, not scored again) advances the re-initialized state as the new segment's \emph{first effective write} ($a:0\to1$). Residual/writer projections are zero-initialized when the adapter is introduced; supplementary Algorithm~S1 gives the execution order. The slot-wise scan recurrence does not establish cross-frame stability: $\rho_t$ in~\eqref{eq:leak} controls direct memory retention, while $\alpha$ in~\eqref{eq:feedback} attenuates the memory-induced state correction; because the write depends on the carried state and memory, their coupled recurrence can still amplify perturbations.

\subsection{Training and Streaming Execution}
The inherited encoder, decoder, state initialization, pose memory, and geometric heads are frozen; the trainable adapter contains 1,278,592 parameters, and a checkpoint audit verifies exact equality of the inherited weights to the pretrained backbone. The loss retains confidence-weighted geometric/pose regression and RGB reconstruction from the existing training implementation:
\begin{equation}
 \mathcal L=\mathcal L_{\mathrm{geom+pose}}^{\mathrm{conf}}+
 \mathcal L_{\mathrm{RGB}}.
\end{equation}
Long input histories establish the recurrent context, while truncated backpropagation detaches scene, pose, and added memory every four supervised recurrent steps. Detachment does not reset their values. Training provenance and the final optimization schedule are given in the supplementary material.

For fixed slot count $L$, scan width $D_s$, and SSM dimension $N$, the selective scan costs $O(LD_sN)$ per frame and the memory occupies $O(LK)$ elements; the extra evaluation in~\eqref{eq:feedback} changes the constant, not the dependence on video length. Section~\ref{sec:results} reports measured totals including encoding, heads, and temporaries.

%% file: sections/04_protocol.tex
\section{Experimental Protocol}
\label{sec:protocol}
Sensor data are recorded first---public benchmark sequences, or traversals captured by the wheelchair platform of Section~\ref{sec:robot_protocol}---and inference replays each recording in temporal order, with one prediction scored per unique image. All front-ends except Spann3R process observations without access to future frames; Spann3R retains its native one-frame look-ahead. The loop-closing back-end of Section~\ref{sec:mapping} operates on accumulated history using the front-end's own predictions. All wheelchair experiments, including the onboard measurements of Section~\ref{sec:device_runtime}, are offline processing of pre-recorded traversals, and statements about speed concern computational throughput relative to the 10~Hz sensor rate under this replay discipline.

\subsection{Datasets, Budgets, and Baselines}
The 500-frame accuracy benchmarks are ScanNet~\cite{dai2017scannet} and KITTI-360~\cite{liao2023kitti360} over five and three sequences; a few Bonn~\cite{palazzolo2019refusion} and TUM RGB-D~\cite{sturm2012tum} scenes measure changing foreground geometry in the supplementary material. The longer trajectory experiment uses KITTI-360 test\_2 and test\_3 at 1,000, 2,000, and nominally 3,000 frames (2,848 valid observations for test\_2 at the last tier; no frames repeated). Within each tier all methods receive the same ordered observations and are averaged equally; all main comparisons use one fixed trained checkpoint, the matched ablations their respective fixed terminal checkpoints.

The comparison includes CUT3R~\cite{wang2025cut3r}, TTT3R~\cite{chen2025ttt3r}, Point3R~\cite{wu2025point3r}, Spann3R~\cite{wang2025spann3r}, ZipMap~\cite{jin2026zipmap}, StreamVGGT~\cite{zhuo2026streamvggt}, STream3R~\cite{lan2026stream3r}, and TTSA3R~\cite{zheng2026ttsa3r}. The first seven are the comparators of the 500-frame and long-trajectory benchmarks; TTSA3R is a training-free comparator on the same released CUT3R checkpoint, evaluated on all five ScanNet scenes and three KITTI-360 blocks. The exact protocol for each method is given in the supplementary material. Mem3R~\cite{liu2026mem3r} and LoGeR~\cite{zhang2026loger} are positioned in Section~\ref{sec:related_state} but not scored: the former replaces the backbone itself and ships no weights, the latter's reported configurations sit above the budget targeted here. CUT3R, TTT3R, and \ours{} use the same reset100 convention: a boundary observation is repeated to initialize the next segment, one prediction per unique image is scored, and segment transforms are composed from predicted cameras---neither ground-truth poses nor per-segment similarity fits enter the stitching. The remaining methods reinitialize their native states, caches, or fast weights every 100 unique observations under the same rigid bridge without segment-wise scale correction; the bridge expression and per-method settings are in the supplementary material.

The budget the design targets is the onboard one---an RTX 4060 with 8~GB, 7.5~GiB allowed as process budget (Section~\ref{sec:device_runtime}), from the platform this work serves; every comparator with a plausible in-budget configuration is evaluated, under a cadence applied to all methods alike.

\subsection{Metrics and Statistical Treatment}
\label{sec:metrics}
A single similarity transform is fitted between the predicted and reference camera centres of each sequence, and ATE is the root-mean-square position error after it. For depth, the scale is estimated once per sequence as the ratio of reference to predicted median camera-$Z$ on fixed reference-valid pixels and is not refitted per frame, reset, or region; AbsRel is the mean relative error under that single scale. Reconstruction accumulates native point maps transformed by the predicted cameras, to which the camera-derived sequence Sim3 is applied without additional ICP; accuracy and completeness are the two directed nearest-neighbour distances and CD is their mean. Per-dataset sample counts and voxel sizes are stated in the supplementary material. All methods share the same reference-valid scoring pixels, voxel origin, and sampling rule.

All accuracy tables and the reported deployment timings use the same kernel configuration (default SDPA dispatch, TF32, non-deterministic execution), fixed checkpoints, and ordered inputs; each method's FP32/BF16 settings are identical between accuracy and deployment runs, and historical resource measurements are separated in the supplement. Where we claim an advantage we test the paired per-sequence differences with an exact one-sided sign test, and we report the median per-sequence relative improvement as the effect size. Each family of comparisons is Holm-corrected for multiplicity and only the corrected value is quoted in this paper. Cohorts of three sequences are reported as counts without a test, since no outcome there can reach significance.

\subsection{Robot-Recorded Cohort and Reference Construction}
\label{sec:robot_protocol}
The recording platform is a powered wheelchair carrying an RS-LiDAR-16, an Intel RealSense D435i, and an RTX 4060-equipped industrial computer. The cohort is ten indoor and three outdoor traversals, recorded on the same platform under the same settings and reference construction; the indoor traversals span several campaigns and are described as one indoor group. Every traversal is decimated to 10~Hz and evaluated at its full retained length rather than truncated to a common budget, so the cohort spans 792 to 4,185 observations. All three evaluated methods receive exactly the same ordered list at $384\times512$ under reset100.

Two properties of the indoor group matter for the mapping study of Section~\ref{sec:mapping}, stated by sequence: I3, I5, I8, I9, and I10 return through their own starting area and so contain an internal revisit a single-session back-end can use, while the rest are single passes; independently, I4--I6 are three sessions of one workspace with shared partial paths and I9--I10 two further path-sharing sessions, so both sets can be fused across sessions. The outdoor traversals are open routes with no usable within-session revisit.

Depth and reconstruction are scored only where the D435i returns a dependable observation. Outdoor geometry routinely exceeds that range, so an outdoor depth or CD value describes the near-field surfaces the sensor actually supports rather than the traversed scene, and is not comparable in absolute terms to the indoor values. On the outdoor traversals we therefore report trajectory error alone, since the LiDAR-anchored trajectory reference is not subject to that range limit, and the near-field outdoor surface and depth values are retained in the evaluation record. This rule follows from the sensor's range rather than from the resulting scores, and applies to every method and every outdoor row alike.

\emph{Reference construction.} The reference camera trajectory comes from the platform's LiDAR stack: dense LiDAR-inertial odometry (FAST-LIO~\cite{xu2021fastlio}) anchored to the optimized keyframe poses of the final LiDAR SLAM map, transformed into the camera frame through the platform's calibrated LiDAR-to-camera extrinsic. The reference depth is the D435i's own recorded stream, and the reference point cloud used by CD back-projects its valid depth observations with the reference poses---the surface reference is what the depth camera observed, placed where the LiDAR trajectory says the camera was, not an independent survey. Reference quantities are used exclusively for scoring: none enters inference, loop verification, or back-end optimization. The supplementary material specifies the anchoring and interpolation, timestamp association, calibration provenance and synchronization limitations, structural quality checks, and construction of the shared multi-session reference frame, including its dependence on candidate initializations from the evaluated methods. The geometric comparisons run on the evaluation host, and Section~\ref{sec:device_runtime} separately measures recorded-input execution on the RTX 4060 system.

%% file: sections/05_results.tex
\section{Evaluation on Public Benchmarks}
\label{sec:results}
This section evaluates \ours{} under the protocol of Section~\ref{sec:protocol}: accumulated geometry at a fixed 500-frame budget, accumulated trajectory accuracy over long streams, the inference cost at which both are obtained, and matched comparisons examining the mixing-operator choice and auxiliary-memory carry-over.
\subsection{Sustained Geometry at 500 Frames}
\begin{figure}[!t]
\centering
\includegraphics[width=\columnwidth]{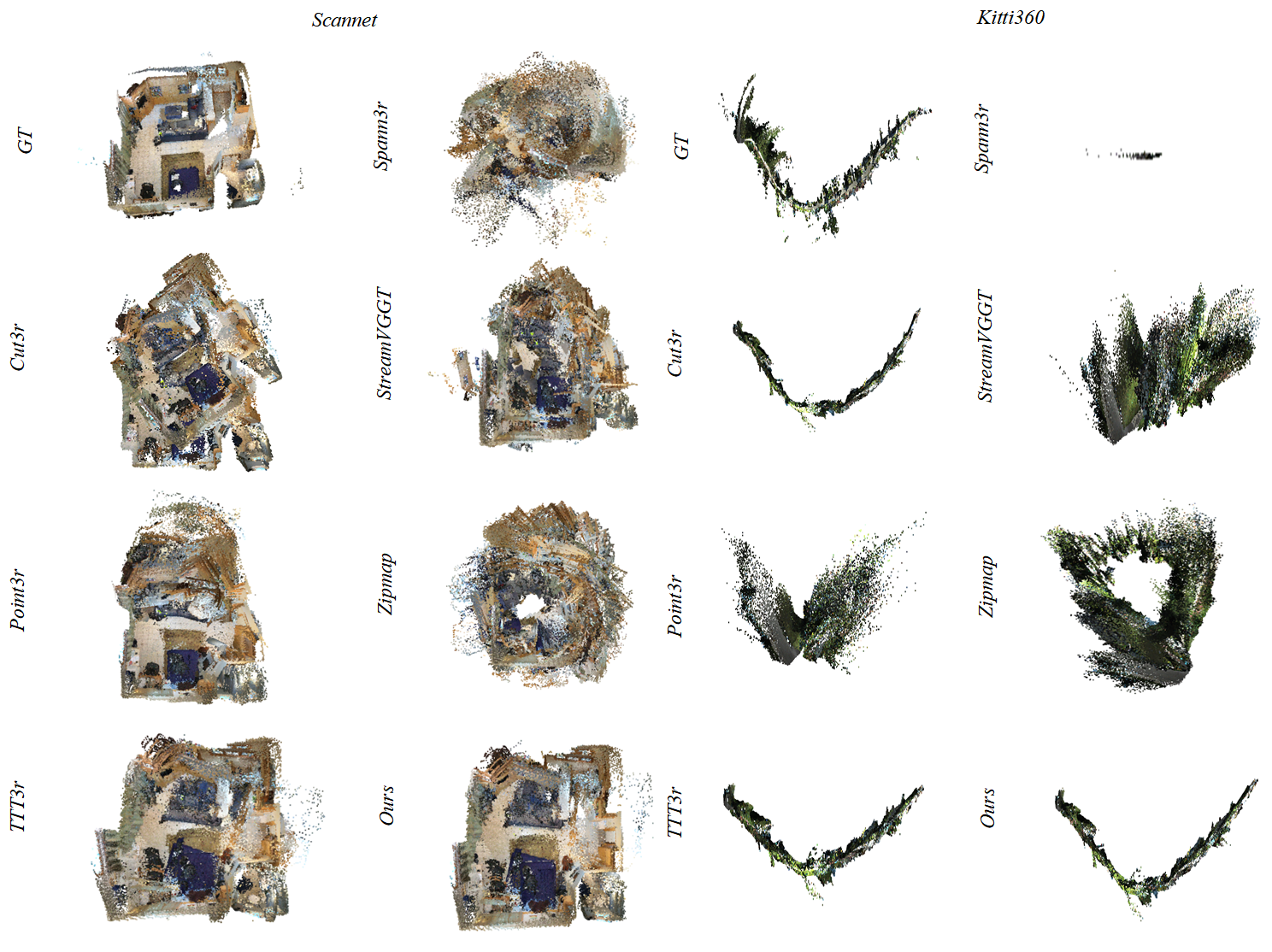}
\caption{500-frame point clouds for the reference (GT) and every evaluated method on ScanNet and KITTI-360; \ours{} stays registered where the cached and spatial-memory comparators fragment.}
\label{fig:clouds500}
\end{figure}

\input{tables/benchmark500_main}
Table~\ref{tab:benchmark500} reports 500-frame geometry on ScanNet and KITTI-360, and Figure~\ref{fig:clouds500} shows the corresponding point clouds. Every row shares one reset cadence, so the comparison isolates the state--memory interface rather than the segmentation schedule. The supplementary no-reset ladder motivates that cadence: on the two KITTI-360 sequences \ours{}'s continuous ATE is 121.45/144.06~m at 1,000/2,000 frames against 25.54/94.88~m at reset100, so reset reduces accumulated error in the tested conditions, without establishing a universally optimal cadence or uniform superiority in continuous operation.

On ScanNet, \ours{} improves on reset-matched TTT3R by 9.6\% in CD, 7.3\% in ATE, and 17.7\% in sequence-scaled AbsRel; KITTI-360 stresses the same interface under fast forward motion and sparse LiDAR support, and the margins grow to 25.1\%, 11.7\%, and 23.9\% (Table~\ref{tab:benchmark500}). TTSA3R~\cite{zheng2026ttsa3r} is retained as a supplementary comparison subject to implementation and protocol limitations. Its evaluated whole-sequence driver yields reset100 mean ATE of 2.071~m on ScanNet and 86.99~m on KITTI-360, but full equivalence to the published protocol has not been established. These values describe the evaluated implementation and are not used to establish superiority over TTSA3R as published; the implementation checks and limitations are detailed in the supplementary material.

Point3R keeps the strongest per-frame depth, consistent with a spatial pointer memory that can be re-read directly, while the compact learned state gives better accumulated pose and surface registration: its CD is 3.7$\times$ ours and its ATE 2.3$\times$ and 3.1$\times$. The attention-cached models are not compared at equal cost: they need 2.3--2.9$\times$ the allocated memory at less than half the throughput (Table~\ref{tab:efficiency}). TTSA3R is excluded from the throughput table because its timed path omits stages the other rows include; the supplementary material times it next to CUT3R under one identical stage list.

Across the cohort, \ours{} attains the lower mean CD, ATE, and AbsRel against both recurrent baselines on ScanNet and KITTI-360, with per-sequence effect sizes and paired tests in the supplementary material.

\subsection{Accumulated Trajectory Accuracy on Long Streams}
\input{tables/long_ate}
Long streams test whether the advantage persists after many recurrent segments. Table~\ref{tab:long} averages the two KITTI test blocks at up to 3,000 frames, with the same checkpoint and reset rule at every tier and no segment-wise reference fitting; under reset100 cross-segment continuity comes from predicted-camera bridging, so this evaluates accumulated geometry under bounded recurrent context rather than uninterrupted memory retention. \ours{} leads at every tier, by 22.2\% over TTT3R at 1,000 frames with less than half the error of reset-matched CUT3R; at the longer tiers the margin over CUT3R narrows to roughly a quarter while remaining the lowest of the three (Table~\ref{tab:long}).

These routes span 0.69--2.67~km, so the absolute errors are best read against distance travelled: normalized, the ATE values are 2.9--5.2\% of route length for \ours{}, against 3.6--5.8\% for TTT3R and 4.9--6.9\% for reset-matched CUT3R. The predictions are monocular under one global scale per kilometre-scale route, so slow scale drift is charged to translation; re-estimating scale per subsequence lowers mean relative translation drift by roughly a third without changing the ordering, and the scale-free rotation drift shows the same ordering, so the margins are not a scale artifact. The front-end is strictly causal, committing each frame once from $144\times512$ crops with no windowed refinement, loop closure, or bundle adjustment; Section~\ref{sec:mapping} measures how much of this open-loop drift a fixed back-end recovers when revisits exist.

\subsection{Inference Cost and the Long-Sequence Operating Point}
\input{tables/efficiency}
Table~\ref{tab:efficiency} includes both adapter evaluations of~\eqref{eq:feedback}. At 500 frames the two recurrent baselines sit within 6.2\% of \ours{} in throughput and 0.3\% in allocated memory; from 1,000 to 2,000 frames \ours{}'s throughput changes by at most 1.4\% with allocated memory unchanged, and it is 2.1$\times$ faster than the two attention-cached comparators while allocating 2.3--2.9$\times$ less. Detailed accounting is in the supplementary material; Section~\ref{sec:device_runtime} measures the consumer GPU directly.

\subsection{Matched Ablations on ScanNet}
\label{sec:scannet_ablation}
\input{tables/scannet_ablation}
Table~\ref{tab:scannet_ablation} compares a matched memory-continuation pair, four freshly initialized mixers, and two inference-time interventions. The memory pair retains the native recurrent states; propagating the added memory reduces CD, ATE, and AbsRel by 1.2\%, 6.8\%, and 3.8\%, respectively.

The four mixers share the frozen CUT3R backbone, memory interface, ordered training inputs, and 1,000-update budget, and their registered mixer parameter counts differ by less than 0.1\%. Alongside pointwise MLP and SS1D, a fixed-kernel cross-slot convolution tests local slot interaction, and lightweight attention tests content-dependent mixing without an SSM recurrence. Both outperform MLP on all three ScanNet metrics, but SS1D remains best: relative to convolution it reduces CD/ATE/AbsRel by 5.9\%/6.7\%/2.9\%, and relative to attention by 4.1\%/3.3\%/4.8\%. These matched-budget results support SS1D over the tested alternatives, not the unique necessity of SSM selectivity. Active capacity and computation are not matched. Architectures, inactive parameters, and training details are specified in the supplementary material.

With fixed weights, the tested feedback settings $\alpha\in\{0,0.25,0.5\}$ stay within 5\% of the evaluated configuration. Full-strength feedback ($\alpha=1$) raises CD/ATE by 74.0\%/43.3\%. Removing decay does not hurt the reported reset100 ATE checks, but raises continuous no-reset ScanNet ATE from 0.6277 to 0.9952~m (58.5\%); its measured benefit is in the tested continuous memory-carry regime. The supplementary material gives the detailed interventions.

%% file: tables/benchmark500_main.tex
\begin{table}[!t]
\centering
\caption{500-frame geometry on ScanNet and KITTI-360 under matched reset100. CD and ATE in meters; depth uses one sequence-wide scale. $^{\dagger}$Spann3R camera recovery; TTSA3R protocol notes are in the supplement.}
\label{tab:benchmark500}
\scriptsize
\setlength{\tabcolsep}{4pt}
\renewcommand{\arraystretch}{0.95}
\begin{tabular*}{\columnwidth}{@{\extracolsep{\fill}}lrrr}
\toprule
Method & CD$\downarrow$ & ATE$\downarrow$ & AbsRel$\downarrow$ \\
\midrule
\multicolumn{4}{@{}l}{\textit{ScanNet~\cite{dai2017scannet} (5 seq.)}} \\
CUT3R & 0.1696 & 0.6540 & 0.05165 \\
TTT3R & 0.1142 & 0.4951 & 0.05670 \\
TTSA3R & 0.8900 & 2.071 & 0.04978 \\
Point3R & 0.3830 & 1.068 & \textbf{0.03164} \\
Spann3R$^{\dagger}$ & 0.9037 & 1.972 & 0.2188 \\
STream3R-$\beta$ & 0.2983 & 1.121 & 0.1753 \\
ZipMap-W1 & 0.3285 & 1.240 & 0.08318 \\
StreamVGGT & 0.1825 & 0.9599 & 0.1685 \\
\ours{} & \textbf{0.1032} & \textbf{0.4589} & 0.04666 \\
\addlinespace
\multicolumn{4}{@{}l}{\textit{KITTI-360~\cite{liao2023kitti360} (3 seq.)}} \\
CUT3R & 2.704 & 11.12 & 0.1917 \\
TTT3R & 2.287 & 8.819 & 0.2472 \\
TTSA3R & 17.98 & 86.99 & 0.19433 \\
Point3R & 6.326 & 24.50 & \textbf{0.1502} \\
Spann3R$^{\dagger}$ & 5.108 & 24.87 & 0.4174 \\
STream3R-$\beta$ & 2.971 & 10.36 & 0.1989 \\
ZipMap-W1 & 6.421 & 16.58 & 0.1720 \\
StreamVGGT & 1.848 & 16.83 & 0.3374 \\
\ours{} & \textbf{1.712} & \textbf{7.788} & 0.1881 \\
\bottomrule
\end{tabular*}
\end{table}

%% file: tables/long_ate.tex
\begin{table}[!t]
\centering
\caption{Long-sequence KITTI-360 ATE (m$\downarrow$), averaged over test\_2 and test\_3 at reset100; tiers are prefixes of the same inputs (the nominal 3,000-frame tier holds 2,848 and 3,000 frames).}
\label{tab:long}
\scriptsize
\setlength{\tabcolsep}{4pt}
\renewcommand{\arraystretch}{1.0}
\begin{tabular*}{\columnwidth}{@{\extracolsep{\fill}}lrrr}
\toprule
Input tier & \ours{} & TTT3R & CUT3R + reset \\
\midrule
1,000 & \textbf{25.54} & 32.81 & 57.23 \\
2,000 & \textbf{94.88} & 105.6 & 125.0 \\
3,000 (2,848 / 3,000) & \textbf{100.6} & 114.4 & 127.4 \\
\bottomrule
\end{tabular*}
\end{table}

%% file: tables/efficiency.tex
\begin{table}[!t]
\centering
\caption{Inference cost on H200 at reset100 under the accuracy-table kernel configuration. The 500-frame block averages one ScanNet and one KITTI-360 input. FPS is the synchronized inference-loop rate (stage list in the supplement); Alloc./Proc.: allocated-tensor and sampled process peaks (GiB).}
\label{tab:efficiency}
\scriptsize
\setlength{\tabcolsep}{4pt}
\renewcommand{\arraystretch}{1.0}
\begin{tabular*}{\columnwidth}{@{\extracolsep{\fill}}lrrr}
\toprule
\multicolumn{4}{c}{\emph{500 frames}} \\
Method & FPS$\uparrow$ & Alloc. & Proc. \\
\midrule
CUT3R & 15.52 & 5.387 & 6.522 \\
TTT3R & 14.02 & 5.388 & 6.849 \\
Point3R & 4.608 & 3.608 & 8.050 \\
Spann3R$^{\dagger}$ & 1.776 & 2.805 & 3.757 \\
STream3R-$\beta$ & 7.036 & 15.70 & 80.57 \\
ZipMap-W1 & 4.404 & 5.832 & 6.852 \\
StreamVGGT & 6.894 & 12.68 & 33.20 \\
\ours{} & 14.61 & 5.404 & 6.854 \\
\midrule
\multicolumn{4}{c}{\emph{1{,}000 frames}} \\
Method & FPS$\uparrow$ & Alloc. & Proc. \\
\midrule
CUT3R & 15.62 & 4.617 & 5.580 \\
TTT3R & 15.59 & 4.616 & 5.740 \\
\ours{} & 14.96 & 4.633 & 5.742 \\
\midrule
\multicolumn{4}{c}{\emph{2{,}000 frames}} \\
Method & FPS$\uparrow$ & Alloc. & Proc. \\
\midrule
CUT3R & 15.61 & 4.618 & 5.506 \\
TTT3R & 15.86 & 4.616 & 5.631 \\
\ours{} & 14.76 & 4.633 & 5.742 \\
\bottomrule
\end{tabular*}
\end{table}

%% file: tables/scannet_ablation.tex
\begin{table}[!t]
\centering
\caption{Matched ablations on the five 500-frame ScanNet sequences at reset100: a memory-continuation pair and four mixers at matched step-1000 checkpoints, plus inference-time feedback and decay interventions (Section~\ref{sec:scannet_ablation}). CD/ATE in meters.}
\label{tab:scannet_ablation}
\scriptsize
\setlength{\tabcolsep}{3pt}
\begin{tabular*}{\columnwidth}{@{\extracolsep{\fill}}lrrr}
\toprule
Configuration & CD$\downarrow$ & ATE$\downarrow$ & AbsRel$\downarrow$ \\
\midrule
\multicolumn{4}{l}{\textit{Memory continuation}} \\
Zero added memory & 0.1021 & 0.4790 & 0.04838 \\
Full added memory & \textbf{0.1009} & \textbf{0.4465} & \textbf{0.04653} \\
\addlinespace
\multicolumn{4}{l}{\textit{New mixer from CUT3R}} \\
Pointwise MLP & 0.1327 & 0.4982 & 0.05864 \\
Cross-slot convolution & 0.1190 & 0.4955 & 0.05176 \\
Lightweight attention & 0.1168 & 0.4784 & 0.05281 \\
Selective SS1D & \textbf{0.1120} & \textbf{0.4625} & \textbf{0.05026} \\
\addlinespace
\multicolumn{4}{l}{\textit{Feedback intervention on the final checkpoint}} \\
No memory increment ($\alpha{=}0$) & 0.1082 & 0.4737 & 0.04693 \\
Attenuated ($\alpha{=}0.25$) & \textbf{0.1028} & \textbf{0.4509} & 0.04682 \\
Evaluated config ($\alpha{=}0.5$) & 0.1032 & 0.4589 & \textbf{0.04666} \\
Undecomposed ($\alpha{=}1$) & 0.1796 & 0.6575 & 0.04702 \\
\addlinespace
\multicolumn{4}{l}{\textit{History-only decay, continuous no-reset rollouts}} \\
With decay ($\rho{<}1$) & \textbf{0.1667} & \textbf{0.6277} & \textbf{0.04424} \\
Without decay ($\rho{=}1$) & 0.1792 & 0.9952 & 0.04656 \\
\addlinespace
\bottomrule
\end{tabular*}
\end{table}

%% file: sections/06_deployment.tex
\section{Onboard Deployment and Robot-Recorded Evaluation}
\label{sec:deployment}
The public benchmarks of Section~\ref{sec:results} fix the geometry--cost operating point on server hardware; this section moves the same checkpoint onto the wheelchair platform of Section~\ref{sec:robot_protocol}: execution cost on its onboard GPU, then geometric quality on its recorded traversals, scored with one alignment per sequence and no global optimization.

\subsection{Execution on a Consumer GPU}
\label{sec:device_runtime}
\input{tables/device_runtime}
Table~\ref{tab:device_runtime} reports execution on an RTX 4060 under the deployment kernel configuration, batch size one at $320\times240$ with reset100 and a 7.5~GiB process budget (PyTorch 2.3.1, CUDA 12.1, driver 575.57.08, i7-12700 host): the front-end in isolation in its top block, the complete coupled system in its bottom block; every run finishes without resolution fallback, non-finite point maps, or dropped frames.

The three recurrent front-ends sit within 5\% of one another in throughput, the \ours{} memory increase over CUT3R is 0.018/0.096~GiB allocated/reserved, and memory does not move with sequence length---the behaviour a constant-capacity state should deliver, measured again at 1,000--2,000 frames in Table~\ref{tab:efficiency}. Against the recording's 10~Hz grid the mean per-frame cost of 85.2~ms leaves headroom inside the 100~ms interval, so the front-end's \emph{average} throughput exceeds the input rate; a fraction of frames still exceeds 100~ms (per-frame p95 of 111~ms). Per Section~\ref{sec:protocol} these are statements about computational capacity on recorded streams, not a live deployment.

The bottom block runs the complete system---same front-end, recorded-bag replay at its 10~Hz timestamps, loop verification, and the deployed incremental back-end (an independent implementation organized after LEMON; Section~\ref{sec:mapping})---as one coupled process, the gate holding at most two packets in flight and applying backpressure rather than dropping. The coupled system commits poses at 9.85~FPS end to end (p50/p95: pose latency against each frame's arrival slot; \emph{drain}: optimizer finish time after the last frame), and the inference path is not the limiter (the forward-only rate matches the front-end block). The supplementary material discusses the accuracy trade-off of this coupled configuration and checks the accuracy of the $320\times240$ deployment setting against the $384\times512$ cohort of Section~\ref{sec:deployment}.

On the onboard device the streaming front-end alone therefore averages above the 10~Hz input rate in offline causal replay, while the complete coupled system averages 9.85~FPS---near real time, about 1.5\% below the grid---with the residual belonging to a back-end designed for offline, whole-sequence optimization rather than to the predictor. Output latency is a separate statement: it is flat at 0.087~s through most of the traversal and rises to a second-scale delay only in the loop-optimization segment (p95 2.83~s)---acceptable for a mapping service that tolerates delayed refinement, not for immediate pose feedback. The supplementary material states the timing origin, plots the per-frame latency with the confirmed-loop instants and back-end busy episodes marked, attributes the delays to their sources, and gives the keyframe-export A/B check.

\subsection{Robot-Recorded Traversals}
\input{tables/robot_forward}
Table~\ref{tab:robot_forward} evaluates the same checkpoint on the cohort of Section~\ref{sec:robot_protocol}, with no retraining and no per-dataset tuning, every traversal scored under the same protocol and reference construction; the sessions I4--I6 additionally enter the multi-session study of Section~\ref{sec:mapping}, while I9 and I10 are evaluated with the deployed incremental back-end and scored on its causal-replay support (keyframe-grid reconstruction), so their CD compares methods within those rows rather than against the other rows. Averaged equally over the ten indoor traversals, \ours{} lowers ATE by 27.9\% and observed-surface CD by 37.5\% relative to reset-matched TTT3R, and by 45.2\% and 44.7\% against the frozen CUT3R backbone under the same schedule, so the adapter improves on the backbone it is attached to and on the strongest baseline at the same time. Per sequence, \ours{} gives the lower ATE, CD, and higher NC on all ten against both baselines---ten of ten each, significant at Holm-corrected $p=0.0078$---though the margins are not uniform: in ATE, I6 separates the two leading methods by less than 2\%, while the smallest margin over CUT3R is 17.1\%. The per-sequence effect sizes are in the supplementary material.

\begin{figure}[!t]
\centering
\includegraphics[width=\columnwidth]{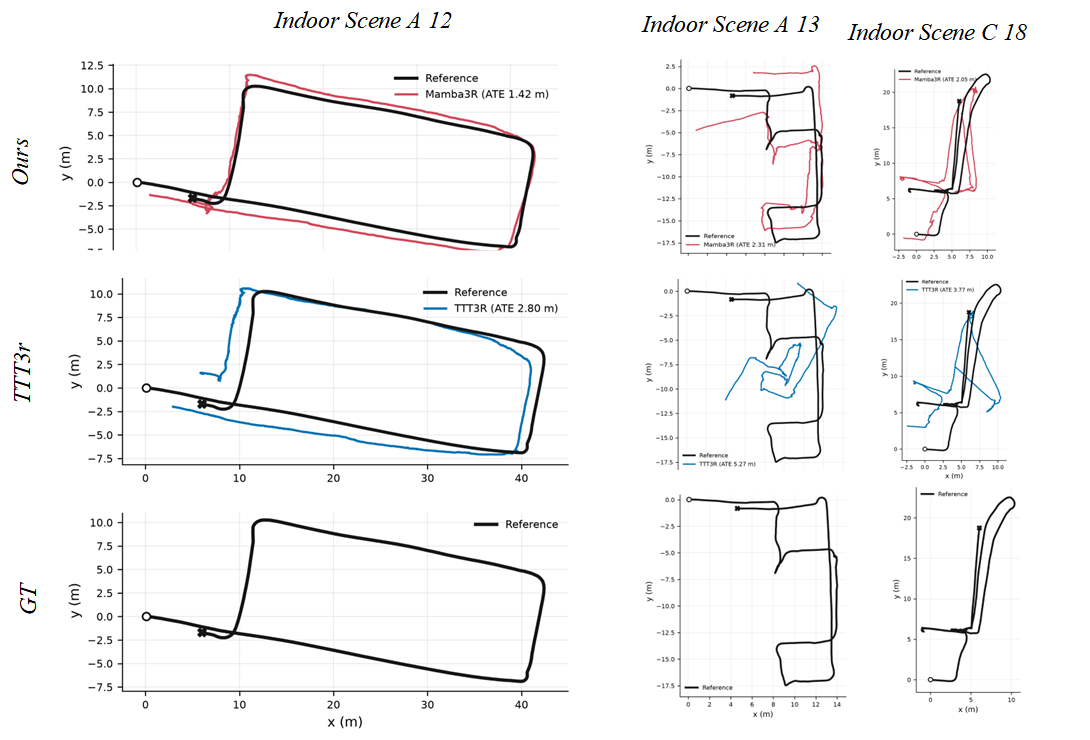}
\caption{Front-end trajectories on I2, I3, and I8: one full-sequence Sim(3) per method, no ICP or segment-wise fit.}
\label{fig:robot_forward}
\end{figure}

\begin{figure}[!t]
\centering
\includegraphics[width=\columnwidth]{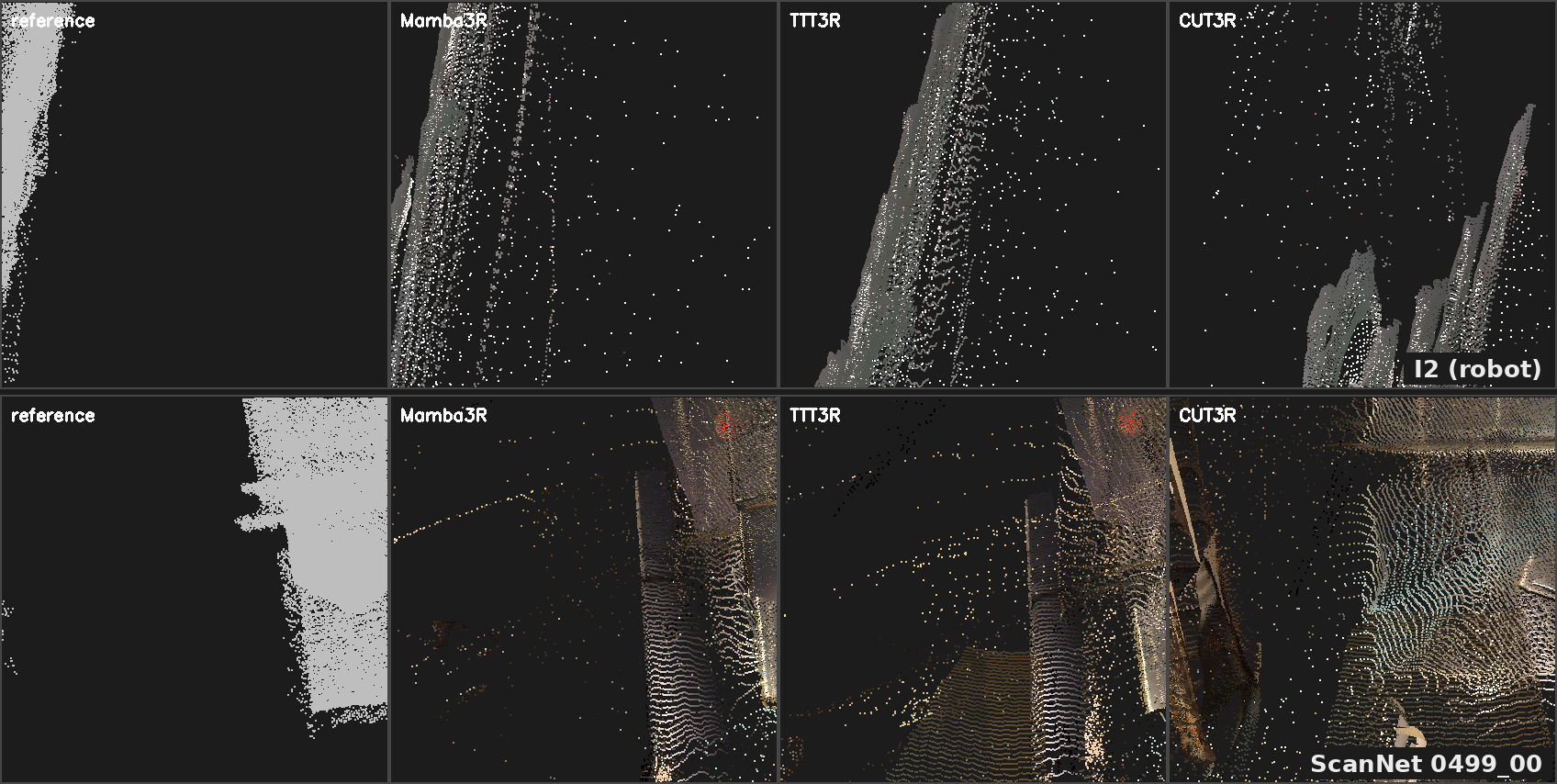}
\caption{Local map detail in two fixed metric crops (top-down slab views): a revisited wall on I2 (top) and ScanNet scene0499\_00 (bottom); each row is one axis-aligned box in the shared reference frame, so position inside a panel is comparable across panels. Crop CDs are in the supplementary material.}
\label{fig:geom_detail}
\end{figure}

Figure~\ref{fig:robot_forward} shows three of these traversals as trajectories: both front-ends recover each route's shape, with errors appearing as an accumulated offset and rotation of the later segments rather than local noise, and the gap is largest where the same space is traversed repeatedly (as on I3, where TTT3R loses the alignment of the repeated passes while \ours{} keeps them registered); neither closes the loop the reference closes, the deficit Section~\ref{sec:mapping} addresses. The three outdoor traversals are reported on trajectory error alone, since beyond the depth camera's dependable range the reference supports no comparable surface or depth measurement: \ours{} has the lowest ATE on all three open routes against both baselines (counts, not tests, for a cohort of three), improving over TTT3R by 21.0--24.9\%. All three still accumulate metres of trajectory error outdoors, the cost of an open traversal to a purely recurrent front-end with no revisit to exploit.

Averages hide where these maps differ, so the supplementary material additionally scores
geometry on depth discontinuities and inside named local regions of the same runs;
Figure~\ref{fig:geom_detail} shows two such regions as fixed metric crops. The edge-restricted
measurement is one-sided in the same direction: \ours{} has the lowest edge CD and the highest
5~cm edge precision on the three self-recorded traversals and on ScanNet scene0499\_00
(supplementary material).

%% file: tables/device_runtime.tex
\begin{table}[!t]
\centering
\caption{Onboard RTX 4060 execution at $320\times240$, reset100, on a recorded 1{,}699-frame traversal under the deployment kernel: the streaming front-end alone (top) and the complete coupled system (bottom). Bold: fastest front-end in the top block; the coupled block is unranked, since its columns trade against one another.}
\label{tab:device_runtime}\label{tab:incremental_device}
\scriptsize
\setlength{\tabcolsep}{4.5pt}
\begin{tabular*}{\columnwidth}{@{\extracolsep{\fill}}lrrrrr}
\toprule
\multicolumn{6}{l}{\textit{Streaming front-end only}} \\
Method & FPS$\uparrow$ & ms/frame$\downarrow$ & Alloc. & Resv. & \\
\midrule
\ours{} & \textbf{11.734} & \textbf{85.2} & 4.670 & 5.131 & \\
TTT3R & 11.399 & 87.7 & 4.650 & 5.092 & \\
CUT3R & 11.262 & 88.8 & 4.652 & 5.035 & \\
\midrule
\multicolumn{6}{l}{\textit{Complete coupled system}} \\
Method & Front$\uparrow$ & p50 (s) & p95 (s) & drain (s) & GPU (GiB) \\
\midrule
CUT3R & 9.993 & 0.107 & 0.756 & 1.017 & 5.242 \\
TTT3R & 9.964 & 0.091 & 0.871 & 0.866 & 5.299 \\
\ours{} & 9.849 & 0.087 & 2.832 & 0.616 & 5.498 \\
\bottomrule
\end{tabular*}
\end{table}

%% file: tables/robot_forward.tex
\begin{table*}[!t]
\centering
\caption{Streaming front-end on the robot-recorded cohort ($384\times512$, reset100, one full-sequence Sim(3), no back-end). CD/ATE in meters; AbsRel at sequence scale; outdoor rows carry trajectory error only (Section~\ref{sec:robot_protocol}). Means cover I1--I10.}
\label{tab:robot_forward}
\scriptsize
\setlength{\tabcolsep}{2.7pt}
\renewcommand{\arraystretch}{0.92}
\begin{tabular*}{\textwidth}{@{\extracolsep{\fill}}lrrrrrrrrrrrrr}
\toprule
& & \multicolumn{3}{c}{ATE$\downarrow$} & \multicolumn{3}{c}{CD$\downarrow$} & \multicolumn{3}{c}{NC$\uparrow$} & \multicolumn{3}{c}{AbsRel$\downarrow$} \\
\cmidrule(lr){3-5}\cmidrule(lr){6-8}\cmidrule(lr){9-11}\cmidrule(lr){12-14}
Seq. & Frames & CUT3R & TTT3R & \ours{} & CUT3R & TTT3R & \ours{} & CUT3R & TTT3R & \ours{} & CUT3R & TTT3R & \ours{} \\
\midrule
\multicolumn{14}{l}{\textit{Indoor}} \\
I1 & 1,989 & 7.017 & 4.141 & \textbf{2.919} & 1.100 & 0.6690 & \textbf{0.3497} & 0.5160 & 0.5092 & \textbf{0.5204} & 0.2872 & \textbf{0.2608} & 0.2817 \\
I2 & 1,591 & 3.626 & 2.796 & \textbf{1.423} & 0.7902 & 0.5225 & \textbf{0.4271} & 0.5400 & 0.5489 & \textbf{0.5574} & \textbf{0.1722} & 0.1765 & 0.2064 \\
I3 & 1,508 & 2.897 & 5.269 & \textbf{2.310} & 0.3512 & 0.8982 & \textbf{0.2967} & 0.5101 & 0.5062 & \textbf{0.5172} & 0.3762 & 0.3862 & \textbf{0.3632} \\
I4 & 1,791 & 4.453 & 2.158 & \textbf{1.867} & 0.6296 & 0.4884 & \textbf{0.3872} & 0.5129 & 0.5204 & \textbf{0.5290} & 0.2717 & 0.2758 & \textbf{0.2386} \\
I5 & 879 & 1.808 & 1.698 & \textbf{1.350} & 0.8740 & 1.138 & \textbf{0.5182} & 0.5316 & 0.5137 & \textbf{0.5453} & 0.2860 & 0.2759 & \textbf{0.1834} \\
I6 & 1,679 & 12.36 & 7.207 & \textbf{7.071} & 3.364 & 1.789 & \textbf{1.505} & 0.5133 & 0.5156 & \textbf{0.5222} & 0.2828 & 0.2643 & \textbf{0.2621} \\
I7 & 792 & 2.576 & 1.837 & \textbf{1.546} & 0.8353 & 1.527 & \textbf{0.6835} & 0.5223 & 0.4990 & \textbf{0.5454} & 0.2992 & 0.2868 & \textbf{0.2624} \\
I8 & 1,640 & 3.619 & 3.769 & \textbf{2.051} & 0.4485 & 0.5298 & \textbf{0.3024} & 0.5128 & 0.5182 & \textbf{0.5214} & 0.3032 & 0.2847 & \textbf{0.2606} \\
I9 & 1,699 & 2.478 & 2.433 & \textbf{2.055} & 0.4681 & 0.4536 & \textbf{0.4345} & 0.5101 & 0.5183 & \textbf{0.5188} & 0.4519 & 0.4296 & \textbf{0.3742} \\
I10 & 1,769 & 3.568 & 2.457 & \textbf{1.750} & 0.8651 & 0.5784 & \textbf{0.4716} & 0.5560 & 0.5717 & \textbf{0.5966} & 0.2181 & 0.2166 & \textbf{0.1978} \\
\addlinespace
\textit{Indoor mean (I1--I10)} & & 4.440 & 3.377 & \textbf{2.434} & 0.9726 & 0.8594 & \textbf{0.5376} & 0.5225 & 0.5221 & \textbf{0.5374} & 0.2948 & 0.2857 & \textbf{0.2630} \\
\addlinespace
\multicolumn{14}{l}{\textit{Outdoor (trajectory error only; see caption)}} \\
O1 & 4,185 & 27.47 & 29.55 & \textbf{23.33} & \multicolumn{9}{c}{---} \\
O2 & 2,087 & 7.423 & 8.310 & \textbf{6.568} & \multicolumn{9}{c}{---} \\
O3 & 2,611 & 6.757 & 8.338 & \textbf{6.262} & \multicolumn{9}{c}{---} \\
\addlinespace
\textit{Outdoor mean (O1--O3)} & & 13.88 & 15.40 & \textbf{12.05} & \multicolumn{9}{c}{---} \\
\bottomrule
\end{tabular*}
\end{table*}

%% file: sections/07_mapping.tex
\section{Front-End Substitution under a Loop-Closing Back-End}
\label{sec:mapping}
Sections~\ref{sec:results} and~\ref{sec:deployment} score the front-end in isolation, so every metre of drift is charged to the recurrent representation; a deployed system detects revisits and redistributes that error, so the relevant question is whether a better front-end still yields a better final map once an optimizer is applied. We therefore hold the back-end fixed within each evaluation block and substitute only the front-end, on the inputs of Section~\ref{sec:robot_protocol} with verifiable revisits: the internal-revisit traversal I8, the three-session indoor fusion set I4--I6, the two-session outdoor set O2--O3, and the two further loop traversals I9--I10 under the deployed incremental back-end. Maps are reported before and after the optimizer runs.

\subsection{The Back-End as a Fixed Instrument}
\label{sec:backend_formulation}
Two instantiations of one loop-closing design serve as the fixed instruments, both independent implementations following LEMON's organization~\cite{wang2026lemon}, neither a contribution of this paper. For I8 and the multi-session blocks it is the \emph{offline continuous-similarity} back-end: an RGB-only similarity pose graph over cached front-end predictions that never re-enters the network, whose nodes are 100-frame submaps linked by odometry priors and verified revisits, with robust optimization yielding a time-interpolated continuous Sim(3) correction applied to poses and point maps. I9--I10 instead use the \emph{deployed incremental} back-end of Table~\ref{tab:backend}, the one the onboard runs of Section~\ref{sec:device_runtime} execute, also fixed across front-ends; its correction is rigid, never rescaling or re-estimating depth. The supplementary material specifies the formulation, thresholds, solver, termination, and implementation checks.

Within each block, admission, verification, submap length, and optimization settings are identical; no reference pose, depth, or scale enters constraint construction or optimization, verification uses predicted geometry with strictly earlier queries, and sessions are never dropped or re-thresholded after scoring. Without a verifiable revisit the front-end trajectory is unchanged, and the offline similarity graph can rescale local depth but not otherwise refine it; Table~\ref{tab:backend} reports depth to expose this limitation.

\subsection{Single Traversal, Multi-Session Fusion, and Shared Alignment}
\label{sec:fusion}
\input{tables/backend_sessions}
The first block of Table~\ref{tab:backend} is the cleanest test of the coupling, because one traversal closes a loop against its own earlier observations and no inter-session registration is involved. The optimizer helps all three front-ends substantially, and the more important observation is what this does to the gap: before optimization \ours{} leads TTT3R by 45.6\% in ATE, afterwards by 4.9\%. The ranking survives, but on a single well-conditioned loop the optimizer recovers most of what a better front-end provided, which is why both columns are reported. The depth column moves the other way on this block, and the mechanism is the correction itself: the offline instrument's continuous Sim(3) field carries a scale component (the verified I8 constraints demand factors of 0.73--0.92 for \ours{}, 0.83--1.28 for TTT3R) that rescales local depth along the route; under one pooled per-session scale the rescaled depths fit slightly worse, so AbsRel rises---least for \ours{}, whose corrections are one-sided---even as trajectory and surface registration improve.

The two later loop traversals repeat the test under the deployed incremental back-end, and I10 shows why trajectory and map quality respond differently to the same correction. The two verified constraints of \ours{} tie one early-pass keyframe (frame 790) to the closing revisit (frames 1640 and 1650); satisfying them closes the seam from 3.4~m to 0.9~m against a 0.2~m reference gap, and ATE falls because the tail's accumulated drift dominates the RMS. The same correction, interpolated between its few anchors, leaves the first half of the route in place (median displacement 0.02~m) but carries 38\% of the trajectory, and every surface observed from it, about a metre sideways. CD scores those accumulated surfaces against the reference everywhere, so walls the front-end had already placed well are re-placed off the reference, and CD rises even as ATE improves. CUT3R, drifted further (seam 7.0~m) with six constraints confined to the same closing segment, shows the signature amplified; TTT3R's ten constraints span 130 frames of the earlier pass, anchoring the bend at many points, and its CD improves with its ATE---so acceptance counts still do not rank front-end quality (TTT3R lands 36\% above \ours{} in final ATE), but they do govern how locally the optimizer can bend the route. A pose graph optimizes camera consistency at verified seams, not surface placement; with few, co-located constraints the trajectory gain does not transfer to the map, which is why Table~\ref{tab:backend} reports both. On I9 only CUT3R confirms a constraint, and it is the one method whose score moves, since a corrected pose history propagates to CD as well even though this back-end never re-estimates depth; \ours{} and TTT3R confirm no constraint, so their optimizers return the front-end trajectory unchanged and their CD repeats the front-end value, leaving \ours{} with the lowest final error and CD and no back-end help.

Multi-session fusion behaves differently, because the sessions were recorded separately and none knows its pose relative to the others. All three front-ends produced graphs in which every requested session became connected. On the three-session indoor set (I4--I6) the optimizer reduces mean ATE and CD for every method, but the gap widens rather than closing: the ATE lead of \ours{} grows from 7.0\% to 20.8\% and its CD lead from 29.4\% to 33.2\%, because almost all improvement concentrates in the one drifting session (7.071 to 2.789~m for \ours{}, 7.207 to 3.490~m for TTT3R) while the already-accurate sessions barely move. CUT3R makes the same point from the other end: it improves by only 11.1\%, because the session dominating its error is the one the optimizer fails to repair (12.4--12.7~m). A front-end whose drift has left too little verifiable overlap cannot be rescued by a back-end that depends on that overlap. The outdoor fusion set, reported on trajectory error alone, shows the weakest coupling: none of the three reached the convergence criterion at the original cap, and with the cap raised tenfold \ours{} converged after 870 cumulative evaluations and TTT3R after 2,818 while CUT3R had still not satisfied it at 3,000. The outdoor mean ATE decreases for all three, and \ours{} retains the lowest final trajectory error (Table~\ref{tab:backend}).

\input{tables/shared_alignment}
Scoring each session against its own reference measures its internal recovery rather than the relative placement of sessions. We therefore additionally evaluate the three-session fusion set under one shared global alignment (Table~\ref{tab:shared_alignment}). The unified reference is constructed by pure-rigid ICP between the observed reference surfaces, initialized from the evaluated methods' cross-session transforms with scale removed. The final registration is selected by a common reference-side inlier criterion and shared by all methods in this check, rather than fitted separately to each method. Because the candidate initializations depend on the evaluated methods, this is a shared-reference consistency check, not validation against a fully method-independent reference. The selected registrations have inlier RMS of approximately 0.10~m at 40--47\% overlap; these residuals are registration diagnostics, not bounds on reference accuracy. The full procedure is given in the supplementary material. For each method, one similarity transform is fitted jointly across all sessions between the optimized camera centres and the registered union reference. Under this shared reference, \ours{} has lower ATE on each session than TTT3R, with 23.4\% lower overall ATE and 17.7\% lower fused-map CD (Table~\ref{tab:shared_alignment}). The outdoor reference surfaces provide only 6\% inlier support, so no shared-alignment claim is made for that set.

%% file: tables/backend_sessions.tex
\begin{table}[!t]
\centering
\caption{Front-end substitution with the back-end fixed within each block: equal-weight means before and after optimization. ATE/CD in meters; AbsRel at per-session scale; the outdoor block carries trajectory error only. Bold: best post-optimization.}
\label{tab:backend}
\scriptsize
\setlength{\tabcolsep}{2.5pt}
\renewcommand{\arraystretch}{0.92}
\begin{tabular*}{\columnwidth}{@{\extracolsep{\fill}}lrrrrrr}
\toprule
& \multicolumn{2}{c}{ATE$\downarrow$} & \multicolumn{2}{c}{CD$\downarrow$} & \multicolumn{2}{c}{AbsRel$\downarrow$} \\
\cmidrule(lr){2-3}\cmidrule(lr){4-5}\cmidrule(lr){6-7}
Method & front & $+$back & front & $+$back & front & $+$back \\
\midrule
\multicolumn{7}{l}{\textit{Single indoor traversal with an internal revisit (I8, 1,640 frames)}} \\
CUT3R & 3.619 & 1.470 & 0.4485 & 0.2858 & 0.3032 & 0.3182 \\
TTT3R & 3.769 & 1.172 & 0.5298 & 0.2748 & 0.2847 & 0.3092 \\
\ours{} & 2.051 & \textbf{1.115} & 0.3024 & \textbf{0.2434} & 0.2606 & \textbf{0.2653} \\
\addlinespace
\multicolumn{7}{l}{\textit{Later-campaign loop traversal I9 (1,699 frames), deployed back-end}} \\
CUT3R & 2.478 & 2.139 & 0.4681 & 0.4890 & 0.4519 & 0.4519 \\
TTT3R & 2.433 & 2.433 & 0.4536 & 0.4536 & 0.4296 & 0.4296 \\
\ours{} & 2.055 & \textbf{2.055} & 0.4345 & \textbf{0.4345} & 0.3742 & \textbf{0.3742} \\
\addlinespace
\multicolumn{7}{l}{\textit{Later-campaign loop traversal I10 (1,769 frames), deployed back-end}} \\
CUT3R & 3.568 & 2.098 & 0.8651 & 0.9441 & 0.2181 & 0.2181 \\
TTT3R & 2.457 & 1.565 & 0.5784 & 0.5399 & 0.2166 & 0.2166 \\
\ours{} & 1.750 & \textbf{1.153} & 0.4716 & \textbf{0.4970} & 0.1978 & \textbf{0.1978} \\
\addlinespace
\multicolumn{7}{l}{\textit{Indoor multi-session fusion (I4, I5, I6; 4,349 frames)}} \\
CUT3R & 6.207 & 5.520 & 1.623 & 1.542 & 0.2802 & 0.2947 \\
TTT3R & 3.688 & 2.309 & 1.138 & 0.7855 & 0.2720 & 0.2955 \\
\ours{} & 3.429 & \textbf{1.828} & 0.8035 & \textbf{0.5243} & 0.2280 & \textbf{0.2322} \\
\addlinespace
\multicolumn{7}{l}{\textit{Outdoor multi-session fusion (O2, O3; 4,698 frames)}} \\
CUT3R & 7.090 & 6.719 & \multicolumn{4}{c}{---} \\
TTT3R & 8.324 & 7.379 & \multicolumn{4}{c}{---} \\
\ours{} & 6.415 & \textbf{6.222} & \multicolumn{4}{c}{---} \\
\bottomrule
\end{tabular*}
\end{table}

%% file: tables/shared_alignment.tex
\begin{table}[!t]
\centering
\caption{Indoor three-session fusion under one shared global alignment: per-session and overall ATE and fused-map CD (meters) under a single Sim(3) to the registered union reference (Section~\ref{sec:fusion}).}
\label{tab:shared_alignment}
\scriptsize
\setlength{\tabcolsep}{4pt}
\begin{tabular*}{\columnwidth}{@{\extracolsep{\fill}}lrrrrr}
\toprule
& \multicolumn{3}{c}{Per-session ATE$\downarrow$} & Overall & Fused \\
\cmidrule(lr){2-4}
Method & I4 & I5 & I6 & ATE$\downarrow$ & CD$\downarrow$ \\
\midrule
TTT3R & 2.634 & 3.871 & 5.015 & 3.705 & 0.419 \\
\ours{} & \textbf{1.798} & \textbf{3.041} & \textbf{3.951} & \textbf{2.839} & \textbf{0.345} \\
\bottomrule
\end{tabular*}
\end{table}

%% file: sections/08_conclusion.tex
\section{Conclusion and Future Work}
We presented \ours{}, a jointly designed state--memory adapter for a frozen recurrent reconstruction backbone: a trainable selective state-space mixer transforms native state features and a fixed-capacity auxiliary memory jointly, explicit read/write paths carry information across frames, and feedback control regulates the memory contribution, while the inherited depth, pose, and point-map interface is preserved.

The contribution is a front-end with fixed added state capacity: under a shared periodic-reset protocol it improves trajectory, accumulated reconstruction, and depth over recurrent baselines on long public and robot-recorded streams at essentially unchanged inference cost, with matched comparisons supporting the mixing-operator choice and the memory carry-over. These results concern accumulated geometry across recurrent segments, not reset-free long-horizon memory generalization; the same checkpoint runs without retuning on a consumer GPU, the regime the design targets.

Handing the same predictions to a fixed loop-closing back-end confirms that the front-end advantage survives global optimization and widens on indoor multi-session fusion.

That validation bounds what comes next. In causal replay the front-end's throughput covers the 10~Hz grid while the coupled system trails by about 1.5\%, a residual belonging to a back-end built for offline whole-sequence optimization. All robot results are offline processing of pre-recorded data, and we do not claim live real-time behaviour for the complete system. Closing that gap---a back-end suited to frame-by-frame monocular streaming, sustaining sensor-rate throughput with bounded output latency---is the next step, together with live calibrated input and having the optimized map correct the compact state in return.